\documentclass[journal]{IEEEtran}

\usepackage{cite}
\usepackage{times}
\usepackage{algorithmic}
\usepackage{array}
\usepackage{mdwmath}
\usepackage{mdwtab}
\usepackage{eqparbox}
\usepackage{cite}
\usepackage{url}
\usepackage{multicol,multirow}
\usepackage{amsmath}
\usepackage{makeidx}
\usepackage{diagbox}
\usepackage{rotating,soul}
\usepackage{wrapfig}
\usepackage{booktabs}
\usepackage{color}
\usepackage{ragged2e}

\usepackage{pifont}
\usepackage{overpic}

\usepackage{amssymb}
\usepackage{colortbl}
\usepackage{bbding}
\usepackage{bm}
\usepackage[colorlinks]{hyperref}

\graphicspath{{./figure/}}
\DeclareGraphicsExtensions{.pdf,.png,.jpg}

\newcommand{\figref}[1]{Fig.~\ref{#1}}
\newcommand{\tabref}[1]{Tab.~\ref{#1}}
\newcommand{\secref}[1]{Sec.~\ref{#1}}

\newcommand{\equref}[1]{Equ. (\ref{#1})}

\newcommand{\mypara}[1]{\noindent {\textbf{#1}}}

\def\ie{\emph{i.e.~}}

\def\sArt{{state-of-the-art~}}

\definecolor{mygray}{gray}{.9}
\definecolor{myblue}{RGB}{223,232,242}

\newcommand{\Yes}{\ding{51}}%

\RequirePackage{silence}
\begin{document}

\title{Spatial Information Guided Convolution for Real-Time RGBD Semantic Segmentation}

\author{\IEEEauthorblockN{Lin-Zhuo Chen, Zheng Lin, Ziqin Wang, 
		Yong-Liang Yang, and Ming-Ming Cheng}\\
	\thanks{LZ Chen(linzhuochen@mail.nankai.edu.cn), Z Lin, MM Cheng (corresponding author, cmm@nankai.edu.cn)
	  are with TKLNDST, College of Computer Science, Nankai University, China.}
	\thanks{Ziqin Wang is with University of Sydney.}
	\thanks{YL Yang is with University of Bath.}
}


\maketitle

\begin{abstract}
3D spatial information is known to be beneficial to the semantic segmentation task.
Most existing methods take 3D spatial data as an additional input,
leading to a two-stream segmentation network that processes
RGB and 3D spatial information separately.
This solution greatly increases the inference time and 
severely limits its scope for real-time applications.
To solve this problem, we propose Spatial information guided 
Convolution (S-Conv), which allows efficient RGB feature and 
3D spatial information integration.
S-Conv is competent to infer the sampling offset of the convolution kernel 
guided by the 3D spatial information, helping the convolutional layer 
adjust the receptive field and adapt to geometric transformations.
S-Conv also incorporates geometric information 
into the feature learning process by generating spatially 
adaptive convolutional weights.
The capability of perceiving geometry is largely enhanced without much
affecting the amount of parameters and computational cost.
%
Based on S-Conv, we further design a semantic segmentation network,
called Spatial information Guided convolutional Network (SGNet),
resulting in real-time inference and \sArt
performance on NYUDv2 and SUNRGBD datasets. 
\end{abstract}


%
\IEEEpeerreviewmaketitle

\section{Introduction}
\IEEEPARstart{W}{ith} the development of 3D sensing technologies,
RGBD data with spatial information (depth, 3D coordinates)
is easily accessible. 
As a result, RGBD semantic segmentation for high-level
scene understanding becomes extremely important, benefiting
a wide range of applications such as automatic driving~\cite{icnet},
SLAM~\cite{bescos2018dynaslam}, and robotics.
Due to the effectiveness of Convolutional Neural Network (CNN)
and additional spatial information, recent advances demonstrate
enhanced performance on indoor scene segmentation tasks
\cite{fcn,deeplab,21SC_WebSeg}.
Nevertheless, there remains a significant challenge caused by the
complexity of the environment and the extra efforts for considering
spatial data, especially for applications that require real-time inference.
\begin{figure}[t]
	\centering
	\begin{overpic}[width=1.\columnwidth,trim=15 0 15 0,clip]{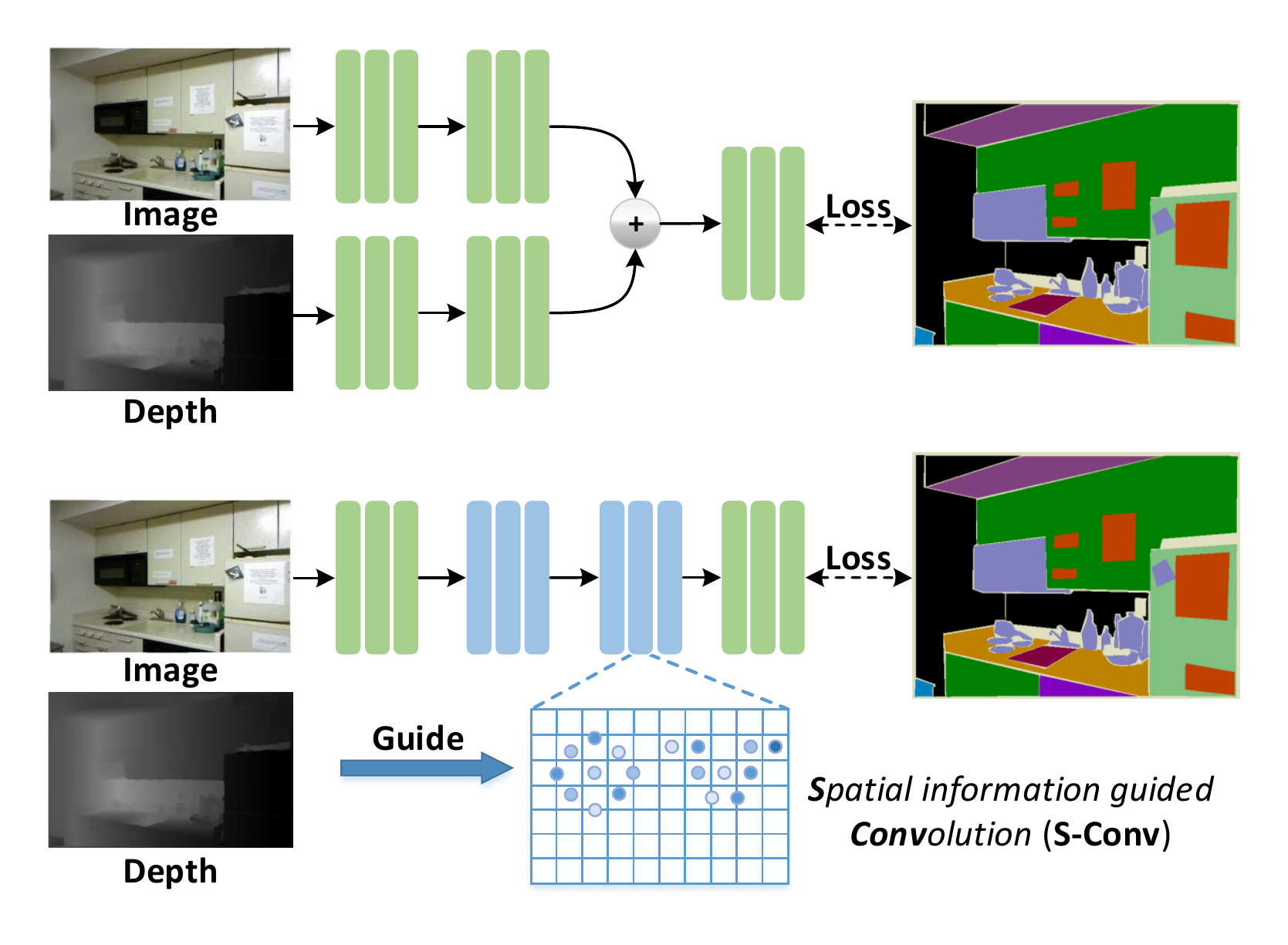}
		\put(46,40){(a)}
		\put(46,0){(b)}
	\end{overpic}
	\caption{\textbf{The network architecture of different multi-modal fusion approaches.} 
			(a) The conventional two-stream structure 
			\cite{rdfnet,eigen2015predicting,ma2017multi,fusenet,wang2016learning}.
			(b) The proposed SGNet.
		It can be seen that the approach in (a)
		largely increases parameter number and
		inference time due to processing spatial information,
		thus less suitable for real-time applications.
		We replace the convolution with our S-Conv in (b)
		where the kernel distribution and weights of the convolution 
		are adaptive to the spatial information.
		S-Conv greatly enhances the spatial awareness of the network
		with few additional parameters and computations,
		thus can efficiently utilize spatial information. Best viewed in color.}
	\label{fig:dataset}
\end{figure}

A common approach treats 3D spatial information as an additional
input, followed by combining
the features of RGB images to fuse multi-modal information
\cite{rdfnet,eigen2015predicting,ma2017multi,fusenet,wang2016learning} (see~\figref{fig:dataset}(a)).
%
%
This approach achieves promising results at the cost of significantly
increasing the parameter number and computational time,
thus being unsuitable for real-time tasks. 
Meanwhile, several works~\cite{fcn, fusenet, gupta2014learning, lstmcf, rdfnet}
encode raw spatial information into three channels
(HHA) composed of horizontal disparity,
height above ground, and norm angle.
%
However, the conversion from raw data to
HHA is also time-consuming~\cite{fusenet}.
It is worth noting that indoor scenes have more complex
spatial relations than outdoor scenes.
This requires a stronger adaptive ability 
of the network to deal with geometric transformations.
%
However, due to the fixed structure of
the convolution kernel, the 2D convolution in the
aforementioned methods cannot well adapt to spatial transformation
and adjust the receptive field inherently, 
limiting the accuracy of semantic segmentation.
Although alleviation can be made by revised pooling operation and
prior data augmentation~\cite{deform, deformablev2},
a better spatially adaptive sampling mechanism for conducting 
convolution is still desirable.

Moreover, the color and texture of objects in indoor scenes
are not always representative \cite{BorjiCVM2019}.
%
Instead, the geometry structure often plays a vital role
in semantic segmentation.
For example, to recognize the fridge and wall,
the geometric structure is the primary cue due to the similar texture.
However, such spatial information is ignored by 2D convolution on RGB data.
%
%
%
The depth-aware convolution~\cite{dcnn} is proposed to address this problem. 
It forces pixels with similar
depths as the center of the kernel to have higher weight than others.
Nevertheless, this prior is handcrafted and may lead to
sub-optimal results.
%


It can be seen that there is a contradiction
between the fixed structure of 2D convolution 
and the varying spatial transformation,
along with the efficiency bottleneck of 
separately processing RGB and spatial data.
To overcome the limitations mentioned above,
we propose a novel operation, called
\emph{\textbf{S}patial information 
guided \textbf{Conv}olution}(S-Conv),
which adaptively changes according to 
the spatial information (see~\figref{fig:dataset}(b)).
%
Specifically, this operation can generate convolution
kernels with different sampling distributions 
adapting to spatial information,
boosting the spatial adaptability and the 
	receptive field regulation of the network.
Furthermore, S-Conv establishes a link
between the convolution weights and the underlying spatial
relationship with their corresponding pixel, incorporating
the geometric information into the convolution weights to
better capture the spatial structure of the scene.
Due to the input of spatial information in S-Conv, 
the scale and spatial transformation of objects can be directly 
analyzed to generate spatially adaptive offsets and weight.
%
%

The proposed S-Conv is light yet flexible and achieves significant
performance improvements with only few additional
parameters  and computation costs, 
making it suitable for real-time applications.
It can be seen as a novel and efficient method for multi-modal fusion task.
Concretely, compared with other two-stream methods, 
we guide the convolution process by utilizing spatial information 
to achieve the purpose of multi-modal fusion. 
It performs better than other methods relying on two-stream network, 
and greatly reduces the amount of parameters and calculation compared 
with two-stream methods, enabling real-time application.
%
We conduct extensive experiments to demonstrate
the effectiveness and efficiency of S-Conv.
We first design the ablation
study and compare S-Conv with two-stream methods,
deformable convolution
\cite{deform, deformablev2} and depth-aware convolution~\cite{dcnn}, 
exhibiting the advantages of S-Conv.
We also verify the applicability of S-Conv to spatial transformations
by testing its influence on different types of spatial
data with depth, HHA and 3D coordinates. We demonstrate that 
spatial information is more suitable to generate offset than 
RGB feature which is used by deformable convolution~\cite{deform, deformablev2}.
Finally, benefiting from the adaptability to spatial transformation
and the effectiveness of perceiving spatial structure,
our network equipped with S-Conv, named 
\textbf{S}patial information \textbf{G}uided convolutional 
\textbf{Net}work (SGNet),
achieves high-quality results with real-time inference on
NYUDv2~\cite{nyud} and SUNRGBD~\cite{sunrgbd, sunrgbd2} datasets.

We highlight our contributions as follows:
\begin{itemize}
	\item
	We propose a novel 
	S-Conv
	operator that
	can adaptively adjust receptive field
	while effectively adapting to spatial transformation, and can perceive
	intricate geometric patterns with low cost.
	%
	\item
	Based on S-Conv, we propose a new 
	SGNet that achieves
	competitive RGBD segmentation performance in real-time on NYUDv2~\cite{nyud} and
	SUNRGBD~\cite{sunrgbd, sunrgbd2} datasets.
\end{itemize}

\section{Related Work}
\subsection{Semantic Segmentation}
The recent advances of semantic segmentation benefit a
lot from the development of convolutional neural network (CNN)
~\cite{imagenet,deep}.
FCN~\cite{fcn} is the pioneer of leveraging CNN for semantic segmentation.
It leads to convincing results and serves as the basic framework for many tasks.
With the research efforts in the field, the recent methods can
be classified into two categories according to the network architecture, 
including atrous convolution based methods~\cite{deeplab, multi, ding2019boundary, shuai2018toward},
and encoder-decoder based methods
~\cite{refinenet, deeplabv3plus, segnet, deconvnet, ding2018context, ding2020semantic}.

\mypara{Atrous convolution:}
The standard approach relies on stride convolutions or poolings to
reduce the output stride of the CNN backbone and 
enables a large receptive field.
However, the resolution of the resulting feature map
is reduced~\cite{deeplab}, and many details are lost.
%
%
One approach exploits atrous convolution to alleviate the
conflict by enhancing the receptive field while
keeping the resolution of the feature map
\cite{deeplab, deeplabv3plus, multi, denseaspp}. 
We use atrous convolution based backbone in the proposed SGNet.

\mypara{Encoder-decoder architecture:}
The other approach utilizes the encoder-decoder structure
\cite{deconvnet, segnet, refinenet, deeplabv3plus, ding2018context, ding2020semantic,Fan2020S4Net},
which learns a decoder to recover the prediction details gradually.
DeconvNet~\cite{deconvnet} employs a series of deconvolutional layers
to produce a high-resolution prediction.
SegNet~\cite{segnet} achieves better results 
by using pooling indices in the encoder to guide 
the recovery process in the decoder.
RefineNet~\cite{refinenet} fuses low-level features
in the encoder with the decoder to refine the prediction. 
~\cite{ding2018context, ding2020semantic} propose a scheme of 
gated sum, which can control the information flow of different scale in the encoder-decoder architecture.
While this method can achieve more precise results, 
it requires longer inference time.

\subsection{RGBD Semantic Segmentation}
How to effectively use the extra geometry information (depth, 3D coordinates)
is the key of RGBD semantic segmentation.
A number of works focus on how to extract more information
from geometry, which is treated as additional
input in~\cite{eigen2015predicting, ma2017multi, fusenet, 
wang2016learning, hu2019acnet}.
Two-stream network is used in \cite{ma2017multi, fusenet, 
wang2016learning, lstmcf, rdfnet}
to process RGB image and geometry information separately,
and combines the two results in the last layer.
These methods achieve promising results at the expense of
doubling the parameters and computational cost.
3D CNNs or 3D KNN graph networks are also used to 
take geometry information into account
\cite{song2017semantic, song2016deep, qi20173d}.
Besides, various deep learning methods on 3D point cloud
\cite{pointnet, pointnet++, chen2019lsanet,spidercnn, spectral_graph_conv,pointcnn} 
are also explored.
However, these methods
cost a lot of memory and are computationally expensive.
Another stream incorporates geometric information into explicit operations.
~\cite{ding2020learning} proposes to perform 3D object detection based on depth-guided convolution, whose weights are location-variant and depth-adaptive.
Cheng et al.~\cite{local} use geometry information to build a feature affinity
matrix acting in average pooling and up-pooling.
Lin et al.~\cite{cascaded} splits the image into different branches based
on geometry information.
Wang et al.~\cite{dcnn} propose Depth-aware CNN, which adds
depth prior to the convolutional weights.
Although it improves feature extraction by convolution, the prior is
handcrafted but not learned from data.
Other approaches, such as multi-task learning
\cite{jiao2019geometry,wang2015towards,hoffman2016learning,
kokkinos2017ubernet,eigen2015predicting,Zhang_2019_CVPR}
or spatial-temporal analysis
\cite{he2017std2p}, are further used to improve segmentation accuracy.
The proposed S-Conv aims to efficiently utilize
spatial information to improve the feature extraction ability.
It can significantly enhance the performance with high efficiency due to using 
only a small amount of parameters.

\subsection{Dynamic structure in CNN}
Using dynamic structure
to deal with varying input of CNN has also been explored.
%
%
Dilation Convolution is used in~\cite{multi, deeplab} to increase the receptive
field size without reducing feature map resolution.
Spatial transformer network~\cite{stn} adapts spatial transformation
by warping feature map.
Dynamic filter~\cite{dynamicfilter} adaptively 
changes its weights according to the input.
Besides, self-attention based methods~\cite{selective, nonlocal, senet, svconv} 
generate attention maps from the intermediate feature map to adjust response
at each location or capture long-range contextual information
adaptively. 
Focusing on the understanding of contextual semantics, shape-variant convolution ~\cite{svconv} confines its contextual region 
by location-variant convolution based on semantic-correlated region.
Some generalizations of convolution from 2D image 
to 3D point cloud are also presented. PointCNN~\cite{pointcnn} is a seminal work that enables CNN on a set of unordered 3D points. 
There are other improvements~\cite{chen2019lsanet,spidercnn,spectral_graph_conv} 
on utilizing neural networks to effectively extract deep features from 3D point sets.
Deformable convolution~\cite{deform,deformablev2}
can generate different distribution with adaptive weights.
Nevertheless, their input is an intermediate feature map rather than
spatial information.
Our work experimentally verifies that better results can be obtained
based on spatial information in \secref{sec:experiments}.


\begin{figure*}[htbp]
	\centering
	\includegraphics[width = \linewidth]{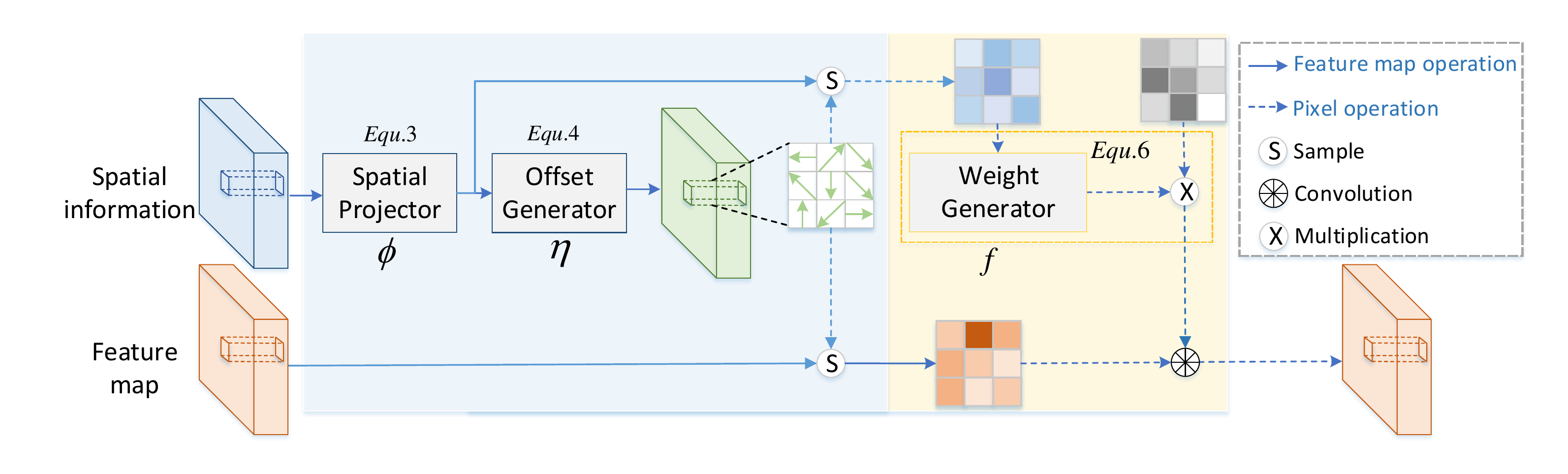}
	\caption{\textbf{The illustration of the Spatial information 
		guided Convolution (S-Conv).} 
		Firstly, the input 3D spatial information is projected 
		by the \textbf{spatial projector}
		to match the input feature map.
		Secondly, the adaptive convolution
		kernel distribution is generated by the \textbf{offset generator}.
		Finally, the projected spatial information is sampled according
		to the kernel distribution and fed into the \textbf{weight generator}
		to generate adaptive convolution weights.}
	\label{fig:sconv}
\end{figure*}
\section{S-Conv and SGNet}
\label{sec:method}
In this section, we first elaborate on the details of 
\emph{\textbf{S}patial information guided \textbf{Conv}olution (S-Conv)},
which is a generalization of conventional 
RGB-based convolution by involving spatial
information in the RGBD scenario.
%
%
%
Then, we discuss the relation between our S-Conv and other approaches.
Finally, we describe the network architecture of 
\textbf{S}patial information \textbf{G}uided convolutional 
\textbf{Net}work (SGNet), which is equipped
with S-Conv for RGBD semantic segmentation.

\subsection{Spatial information guided Convolution}
For completeness, we first review the conventional convolution operation.
We use $\mathbf{A}_i(\mathbf{j}), 
\mathbf{A} \in \mathbb{R}^{c \times h \times w}$ 
to denote a tensor, where
$i$ is the index corresponding to the first dimension, and $\mathbf{j \in \mathbb{R}}^2$
indicates the two indices for the second and third dimensions.
Non-scalar values are highlighted in bold for convenience.

For an input feature map $\mathbf{F}\in \mathbb{R}^{c\times h \times w}$.
We describe it in 2D for simplicity, thus 
we note $\mathbf{X}$ as input feature map.
$\mathbf{X} \in \mathbb{R}^{1 \times h \times w}$.
Note that it is straightforward
to extend to the 3D case.
The conventional convolution applied on $\mathbf{X}$ to 
get $\mathbf{Y}$ can be formulated as the following:
\begin{equation}
\mathbf{Y}(\mathbf{p})=\sum_{i=1}^{K} \mathbf{W}_{i} 
\cdot \mathbf{X}(\mathbf{p}+\mathbf{d}_i),
\label{equ:equ1}
\end{equation}
where $\mathbf{W} \in \mathbb{R}^{K}$
represents the weight of convolution kernel with kernel 
size $k_h \times k_w$, and $K = k_h \times k_w$. $\mathbf{p} \in \mathbb{R}^2$
is the 2D convolution center, $\mathbf{d} \in \mathbb{R}^{K \times 2}$
denotes the kernel distribution around $\mathbf{p}$. For $3 \times 3$
convolution, $d$ is defined as \equref{equ:equ2}:

\begin{equation}
\mathbf{d} = \{[-1, -1], [-1, 0], ... , [0, 1], [1, 1]\}.
\label{equ:equ2}
\end{equation}
From the above equation, we can see that the convolution kernel 
is constant over $\mathbf{X}$. In other words,
$\mathbf{W}$ and $\mathbf{d}$ are fixed, meaning the convolution 
is location-invariant and spatially-agnostic.

\begin{figure}[t]
	\centering
	\vspace{-30pt}
	\includegraphics[width = 1.\columnwidth, trim=0 0 0 0,clip]{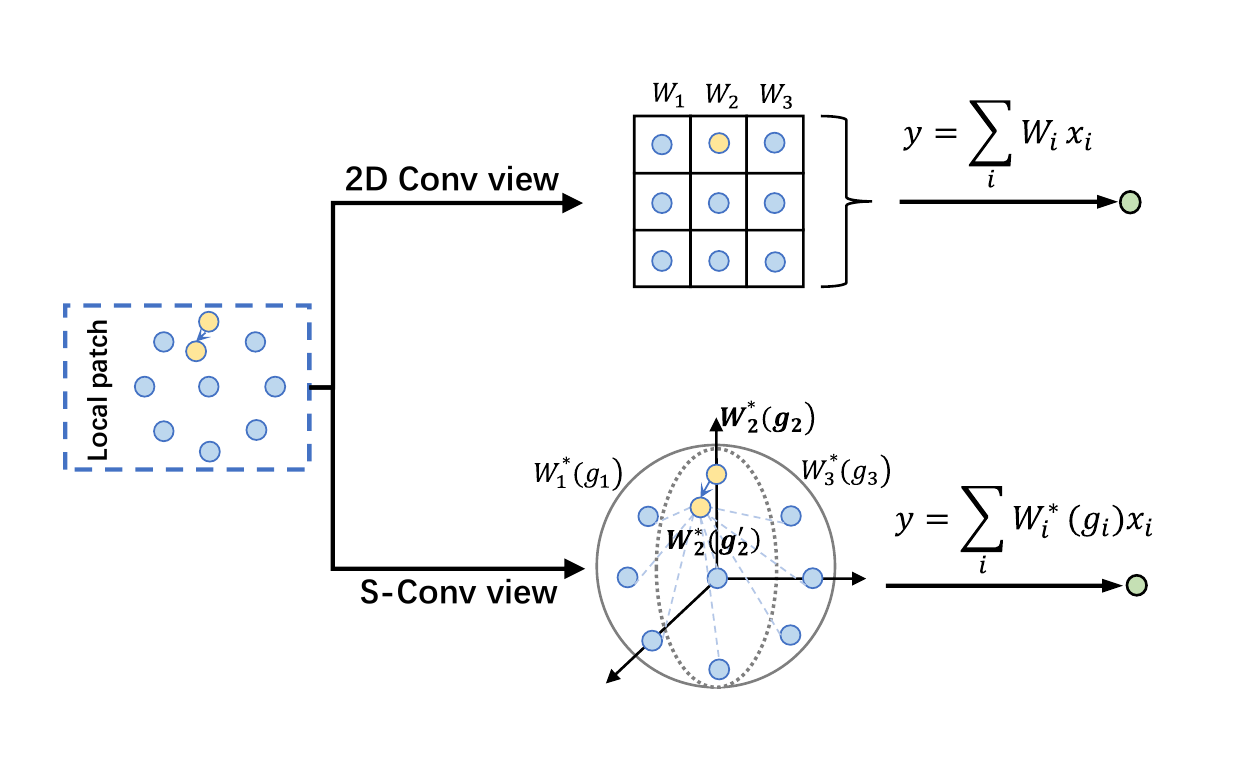}
	\vspace{-30pt}
	\caption{
		\textbf{The illustration of weights $W$ in 2D convolution and $W^*$ in S-Conv.}
		The yellow dot indicates the point whose spatial position changes along the arrow.
		Illustration of 2D convolution is on the top, and S-Conv is on the bottom.
		The conventional 2D convolution operation orderly places local
		points in a regular grid with fixed weights, while ignoring the spatial information.
		We can see that the spatial position variation of the yellow point 
		can not be reflected in the weight.
		Our S-Conv can be regarded as placing a local patch
		into a weight space, which is generated by the spatial guidance of that patch.
		Hence the weight of each point establishes a link with its spatial location,
		effectively capturing the spatial variation of the local patch.
		The spatial relationship between the yellow point and other points
		can be reflected in the adaptive weights.}
	\label{fig:revisit}
\end{figure}
\begin{figure*}[t]
	\centering
	\includegraphics[width = \linewidth, trim=20 20 20 20,clip]{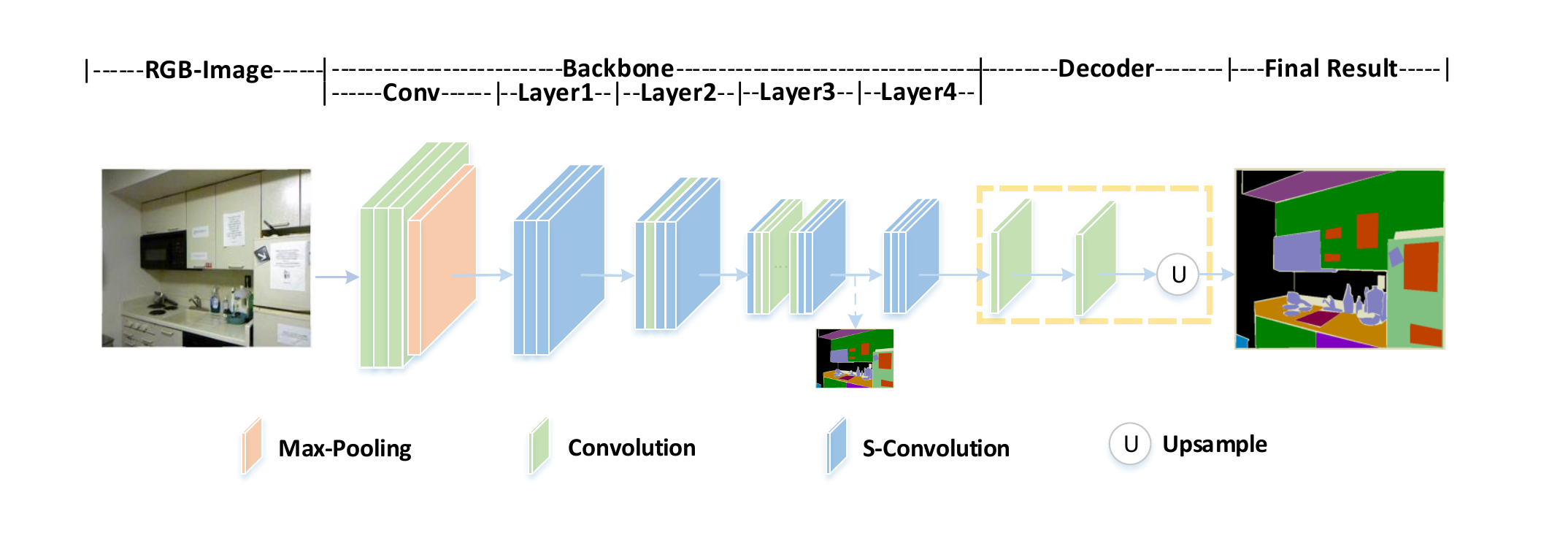}
	\vspace{-20pt}
	\caption{\textbf{The network architecture of SGNet equipped with S-Conv for 
	RGBD semantic segmentation.} 
	The SGNet consists
		of a backbone network and a decoder. 
		The deep supervision is added between layer 3 and layer 4 to 
		improve network optimization.}
	\label{fig:network}
\end{figure*}
In the RGBD context, we want to involve 3D spatial
information efficiently by using adaptive convolution kernels.
We first generate the offset according to the spatial
information, then use the spatial
information corresponding to the given offset
to generate new spatially adaptive weights.
Our S-Conv requires two inputs. One is the feature map $\mathbf{X}$
which is the same as conventional convolution.
The other is the spatial information $\mathbf{S}
\in \mathbb{R}^{c' \times h \times w}$.
In practice, $\mathbf{S}$ can be HHA ($c'=3$), 3D coordinates ($c'=3$), or depth ($c'=1$).
The method of encoding depth into 3D coordinates and HHA is the same as~\cite{qi20173d}.
Note that the input spatial information is not included in the feature map.
%

As the first step of S-Conv, we project the input spatial information into
a high-dimensional feature space, which can be expressed as:

\begin{equation}
\mathbf{S'} = \phi(\mathbf{S}),
\label{equ:equ3}
\end{equation}
where $\phi$ is a spatial transformation function, and
$\mathbf{S}' \in \mathbb{R}^{64 \times h \times w}$,
which has a higher dimension than $\mathbf{S}$.

Then, we take the transformed spatial information $\mathbf{S'}$ into consideration,
perceive its geometric structure, and generate the distribution 
(offset of pixel coordinate in $x-$ and $y-$axis)
of convolution kernels at different $\mathbf{p}$.
This processes can be expressed as:
\begin{equation}
\mathbf{\Delta d} =\eta(\mathbf{S'}),
\label{equ:equ4}
\end{equation}
where $\mathbf{\Delta d} \in \mathbb{R}^{K\times h' \times w' \times 2}$,
For the sake of simplicity, 
we do not show the reshaping process of $\mathbf{\Delta d}$ in \equref{equ:equ4}.
$\mathbf{\Delta d} \in \mathbb{R}^{2K\times h' \times w'}$ before reshaping.
$h', w'$ represent the feature map size after convolution. 
$K = k_h \times k_w$, which $k_h$ and $k_h$ are the kernel size. 
For $3 \times 3$ convolution, 
$\mathbf{\Delta d} \in \mathbb{R}^{9\times h' \times w' \times 2}.$
$\eta$ is a non-linear function which can be implemented
by a series of convolutions. 
%

After generating the distribution of kernel for each possible
$\mathbf{p}$ using $\mathbf{\Delta d(\mathbf{p})}$, 
we boost its feature extraction
ability by establishing the link between the geometric
structure and the convolution weight.
Due to the shifting of convolution kernel in \equref{equ:equ4}, 
the corresponding depth information of the convolution kernel 
has also changed.
We want to collect the depth information corresponding to the convolution kernel after shifting for generating spatially adaptive weight.
More specifically, we sample the geometric information of the
pixels corresponding to the convolution kernel after shifting:
%
\begin{equation}
\mathbf{S}^*(\mathbf{p}) = \{\mathbf{S'}(\mathbf{p} +\mathbf{d}_{i} +
\Delta\mathbf{d}_{i}(\mathbf{p})) |_{ i = 1, 2, ..., K}\},
\end{equation}
where $\mathbf{\Delta d}(\mathbf{p})$ is the spatial
distribution of convolution kernels at $\mathbf{p}$.
$\mathbf{S^*}(\mathbf{p}) \in \mathbb{R}^{64K}$
is the spatial information corresponding to the feature map
of the convolution kernel centered on $\mathbf{p}$
after transformation.

Finally, we generate convolution weights according to the 
final spatial information as the following:
\begin{equation}
\mathbf{W}^{*}(\mathbf{p})=\sigma(f(\mathbf{S}^*(\mathbf{p}))) \cdot \mathbf{W},
\label{equ:equ6}
\end{equation}
where $f$ is a non-linear function that can be implemented
as a series of fully connected layers with non-linear activation function,
$\sigma$ is sigmoid function, $\cdot$ is element-wise product,
$\mathbf{W} \in \mathbb{R}^{K}$ indicates the convolution
weights, which can be updated by the gradient descent algorithm.
$\mathbf{W}^{*}(\mathbf{p}) \in \mathbb{R}^{K}$
denotes the spatially adaptive weights for convolution after shifting
centered at $\mathbf{p}$.

Overall, our generalized S-Conv is formulated as:
\begin{equation}
\mathbf{Y}(\mathbf{p})=\sum_{i=1}^{K}  
\mathbf{W}_{i}^{*}(\mathbf{p}) \cdot \mathbf{X}(\mathbf{p}+\mathbf{d}_{i} +
\Delta\mathbf{d}_{i}(\mathbf{p})).
\label{equ:equ7}
\end{equation}
We can see that $\mathbf{W}_{i}^{*}(\mathbf{p})$ establishes the correlation
between spatial information and convolution weights.
Moreover, convolution kernel distribution is also relevant to the spatial 
information through $\Delta\mathbf{d}$.
Note that $\mathbf{W}_{i}^{*}(\mathbf{p})$ and
$ \Delta\mathbf{d}_{i}(\mathbf{p})$ are not constant, meaning the
generalized convolution is adaptive to different $\mathbf{p}$.
%
%
Also, as $\Delta\mathbf{d}$ is typically fractional,
we use bilinear interpolation to compute $\mathbf{X}(\mathbf{p}+\mathbf{d}_{i} +
\Delta\mathbf{d}_{i}(\mathbf{p}))$ as in~\cite{deform, stn}.
The main formulae discussed above are labeled in \figref{fig:sconv}.

\subsection{Relation to other approaches}
2D convolution is the special case of the proposed
S-Conv without geometry information.
Specifically, without geometry information, if we remove the $\mathbf{W}_{i}^{*}(\mathbf{p})$ and $\Delta\mathbf{d}_{i}(\mathbf{p})$ which are generated by geometry
information in \equref{equ:equ7}, this process degenerates to 2D convolution.
%
%
While for the RGBD case, our S-Conv can extract feature at 
the point level and is not limited to the discrete grid by 
introducing spatially adaptive weights as shown in \figref{fig:revisit}. 
Deformable convolution~\cite{deform, deformablev2} also alleviates
this problem by generating different distribution weights.
Nevertheless, their distributions are inferred from 2D feature maps instead
of 3D spatial information as in our case.
%
We will verify through experiments that our method achieves
better results than deformable convolution~\cite{deform, deformablev2}.
Compared with shape-variant (SV) convolution~\cite{svconv},
SV convolution confines its contextual region by 
location-variant convolution based on semantic-correlated region. It 
implements a location-variant convolution operator whose weights 
are location-variant and generated by feature map, 
focusing on the understanding of contextual semantics. 
Our S-Conv utilizes depth map rather than feature map 
to generate spatially adaptive offsets and weights.
The weights and offset of S-Conv are defined by spatial information (depth map). This
helps the convolutional layer to adjust the receptive field and adapt 
to geometric transformation according to the spatial information.
Compared with the 3D KNN graph-based method, our S-Conv
selects neighboring pixels adaptively instead of using the KNN graph,
which is not flexible and computationally expensive.
%
\subsection{SGNet architecture}
Our semantic segmentation network, called SGNet, is equipped with S-Conv
and consists of a backbone and decoder.
The structure of SGNet is illustrated in \figref{fig:network}.
%
We use ResNet101~\cite{resnet} as our backbone,
and replace the first and the last two conventional convolutions 
($3 \times 3$ filter)
of each layer with our S-Conv.
We add a series of convolutions to extract the feature further
and then use bilinear up-sampling to produce the final segmentation
probability map, which corresponds to the decoder part of the SGNet.
The $\phi$ in \equref{equ:equ3} is implemented as three
$3 \times 3$ convolution layers, 
\ie Conv(3, 64) - Conv(64, 64) - Conv(64, 64) with 
non-linear activation function.
The $\eta$ in \equref{equ:equ4} and the $f$ in \equref{equ:equ6}
are implemented as single convolution layer and two fully connected
layers separately. 
The S-Conv implementation is modified from deformable convolution~\cite{deformablev2,deform}.
We add deep supervision between layer 3 and layer 4 to improve
the network optimization capability, which is the same as PSPNet~\cite{psp}.

\section{Experiments}
\label{sec:experiments}
 In this section, we first validate the performance of S-Conv by analyzing its usage in 
different layers; conducting ablation study/comparison with its alternatives; 
evaluating results of using different input information
to generate offset; and testing inference speed.
Then we compare our SGNet equipped with S-Conv with other \sArt semantic segmentation methods
on NYUDV2 and SUNRGBD datasets.
Finally, we visualize the depth adaptive receptive field in each layer 
and the segmentation results, demonstrating that the proposed S-Conv can well exploit spatial information.
\begin{figure*}[t]
	\centering
	\includegraphics[width = 2.0\columnwidth]{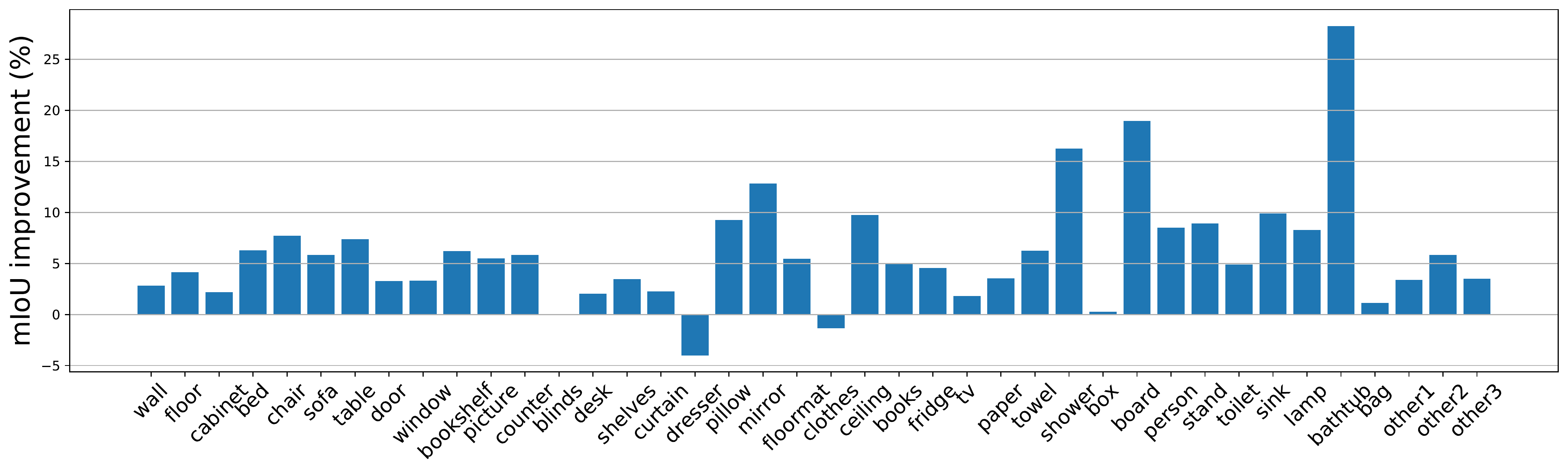}\\
	\vspace{-15pt}
	\caption{\textbf{Per-category IoU improvement of S-Conv on NYUDv2 dataset.}}
	\label{fig:catgory}
\end{figure*}

\begin{table*}[t]\normalsize
	\centering
	\caption{\textbf{The results of replacing convolution (of $3 \times 3$ filter)
			of different layers with S-Conv on NYUDv2 dataset.}
		``layerx\_y" means the $3 \times 3$ convolution of y-th residual block 
		in x-th layer. 
	}\label{tab:ablation1} \vspace{-6pt}
	\renewcommand\tabcolsep{4pt}
	\begin{tabular}{c ||c c c c c c c |c  c c}\toprule[1.5pt]		
		S-Conv & layer3\_0 & layer3\_1 & layer3\_2 & layer3\_20 & layer3\_21 & layer3\_22 & other layers & mIoU(\%) &param(M)&FPS\\ \hline \hline
		&           &          &          &          &          &          &          &43.0         &56.8&34\\
		&\checkmark &          &          &          &          &          &          &47.0         &56.9&34\\
		Baseline           &\checkmark &\checkmark&\checkmark&          &          &          &          &46.6        &57.2&33\\
		(ResNet101)        &           &          &          &\checkmark&\checkmark&\checkmark&          &46.5         &57.2&33\\
		&\checkmark &          &          &          &\checkmark&\checkmark&          &47.8         &57.2&33\\
		&\checkmark &          &          &          &\checkmark&\checkmark&\checkmark&\textbf{49.0}&58.3&26\\
		\bottomrule[1.5pt]
	\end{tabular}
\end{table*}

\begin{table}[ht]\normalsize
	\centering
	\renewcommand\tabcolsep{5.5pt}
	\caption{\textbf{The results of replacing convolution (of $3 \times 3$ filter)
		of different layers with S-Conv on NYUDv2 dataset.}
	}\label{tab:ablation_layer} \vspace{-6pt}
	\begin{tabular}{c c c c |c c}	\toprule[1.5pt]	
		layer1 & layer2 & layer3 & layer4 & mIoU(\%) & param(M)\\ \hline	\hline
		           &          &          &          &43.0       & 56.8\\
    \checkmark &          &          &          &46.5       & 57.2\\
		\checkmark &\checkmark&          &          &47.0       & 57.5\\
	  \checkmark &\checkmark&\checkmark&          &48.8       & 57.9\\
		\checkmark &\checkmark&\checkmark&\checkmark&\textbf{49.0}&58.3\\
		\bottomrule[1.5pt]
	\end{tabular}
\end{table}

\mypara{Datasets and metrics:} We evaluate the performance
of S-Conv operator and SGNet segmentation method on public datasets:
\begin{itemize}
	\item NYUDv2~\cite{nyud}:  This dataset has
	1449 RGB images with corresponding depth maps and
	pixel-wise labels. 795 images are used for training,
	while 654 images are used for testing as in~\cite{split}.
	The 40-class settings are used for experiments.
	\item SUN-RGBD~\cite{sunrgbd, sunrgbd2}: This dataset
	contains 10335 RGBD images with semantic labels organized in 37 categories.
	5285 images are used for training, and 5050 images are used for testing.
	
	%
\end{itemize}

We use three common metrics for evaluation, including pixel accuracy (Acc),
mean accuracy (mAcc), and mean intersection over union (mIoU). 
The three metrics are defined as the following:

\begin{equation}
\begin{aligned}
\centering
&Acc = \sum_i \frac{p_{ii}}{g},\\
&mAcc = \frac{1}{p_c} \sum_i \frac{p_{ii}}{g_i}, \\ 	
mIoU = &\frac{1}{p_c} \sum_i \frac{p_{ii}}{g_i + \sum_j p_{ji} - p_{ii}},
\end{aligned}
\end{equation}
where $p_{ij}$ is the amount of pixels which 
are predicted as class $j$ with ground truth $i$, 
$p_c$ is the number of classes, and $g_i$ is the number of pixels whose ground truth class is $i$. 
$g = \sum_{i} g_i$ is the number of pixels.
The depth map is used as the default 
format of spatial information unless specified otherwise.

\mypara{Implementation details:} We use dilated ResNet101~\cite{resnet}
pretrained on ImageNet~\cite{imagenet-c} as the backbone 
network for feature extraction following~\cite{deeplab}, 
and the output stride is 16 by default.
The whole system is implemented based on PyTorch.
The SGD optimizer is adopted for training with the same learning rate schedule (``poly" policy) 
as~\cite{deeplab, deeplabv3plus},
where the initial learning rate is 5e-3 for ablation study,
8e-3 for NYUDv2 and 1e-3 for SUNRGBD,
and the weight decay is 5e-4. This learning policy updates the learning
rate for every 40 epochs for NYUDv2 and ablation study 
and 10 epochs for SUNRGBD.
We use ReLU activation function, and the batch size is 8.
Following~\cite{rdfnet}, we employ general data augmentation
strategies, including random scaling, random cropping, and random flipping.
The crop size is $480 \times 640$.
During testing, we down-sample the image to the training crop size ($480 \times 640$),
and its prediction map is upsampled to the original size.
We use cross-entropy loss in both datasets, and reweight~\cite{jiang2018rednet}
training loss of each class in SUNRGBD due to
its extremely unbalanced label distribution.
We train the network by 480 epochs for the NYUDv2 dataset and
200 epochs for the SUNRGBD dataset on two NVIDIA 1080Ti GPUs.

\subsection{Analysis of S-Conv}

We design ablation studies on NYUDv2~\cite{nyud} dataset.
The ResNet101 with a simple decoder and deep supervision is used as the baseline.

\mypara{Replace convolution with S-Conv}: We evaluate the effectiveness
of S-Conv by replacing the conventional convolution
(of $3 \times 3$ filter) in different layers.
We first replace convolution in layer 3,
then extend the explored rules to other layers. The FPS (Frames per Second) is
tested on NVIDIA 1080Ti with input image size $480 \times 640$.
The results are shown in \tabref{tab:ablation1}.

\begin{table}[t]\normalsize
	\centering
	\caption{\textbf{Ablation study of SGNet on NYUDv2~\cite{nyud} dataset.}
		OG: Offset generator of S-Conv, WG: Weight generator of S-Conv,
		SP: Spatial projection of S-Conv. }
	\label{tab:ablation2}	\vspace{-7pt}
	\renewcommand\tabcolsep{10.5pt}
	\begin{tabular}{l || c c c }	\toprule[1.5pt]
		Model		       &Acc & mAcc &mIoU\\ \hline \hline
		Baseline 		   &72.1  &54.6 &43.0\\

		Baseline+OG	       &73.9 &58.2&46.3\\
		Baseline+SP+OG     &75.2 &60.0&48.4\\
		Baseline+SP+WG     &74.5 &58.4&46.8\\
		Baseline+SP+OG+WG  &\textbf{75.5}&\textbf{60.9}&\textbf{49.0}\\
		\bottomrule[1.5pt]
	\end{tabular}
\end{table}

\begin{table}[htbp]\normalsize
	\centering
	\renewcommand\tabcolsep{8.0pt}
	\caption{\textbf{The comparison results on NYUDv2 test dataset.}
	    DCV2: Deformable
		Convolution V2~\cite{deformablev2},
		DAC: Depth-aware Convolution~\cite{dcnn}, 
		SP: Spatial projector in S-Conv, WG: Weight generator in S-Conv.
	}
	\label{tab:promote} \vspace{-7pt}
	\begin{tabular}{l ||c c c}
		\toprule[1.5pt]
		Model						&Acc & mAcc & mIoU	 \\
		\hline
		\hline
		Baseline 			       &72.1  &54.6 &43.0\\
		Baseline+DCV2              &73.0 &56.1&44.5\\
		Baseline+HHANet            &73.5  &56.8  &45.4 \\
		Baseline+DAC               &73.8  & 57.1 &45.4 \\
		Baseline+HHANet+DCV2 &74.3  &58.4  &47.0   \\
		Baseline+DAC+DCV2    &74.5  &58.3  &46.5   \\
		Baseline+SP+WG             &74.5  & 58.4 &46.8 \\
		Baseline+S-Conv(SGNet)     &\textbf{75.5}  &\textbf{60.9} &\textbf{49.0}\\
		\bottomrule[1.5pt]
	\end{tabular}
\end{table}

\begin{table}[t]\normalsize
	\centering
	\renewcommand\tabcolsep{14pt}
	\caption{\textbf{Comparison of using different types of spatial 
	information on NYUDv2 Dataset.}}
	\label{tab:ablation3} \vspace{-7pt}
	\begin{tabular}{c || c c c }
		\toprule[1.5pt]
		Information &  Acc 		& mAcc 	& mIoU \\
		\hline
		\hline
		Depth 		&75.5 &	60.9 &\textbf{49.0}    \\
		RGB Feature 	&73.9 &58.5 &46.4 \\
		HHA         &\textbf{75.7} &60.8& 48.9\\
		Coordinates 	&75.3&\textbf{61.2}&48.5  \\
		\bottomrule[1.5pt]
	\end{tabular}
\end{table}

\begin{table}[ht]\normalsize
	\centering
	\caption{\textbf{Inference Speed test of SGNet with input image
	    size $480 \times 640$.}
		OG: Offset generator of S-Conv, \dag : without
		applying generated location-variant weight and offset in SGNet,
		HHANet: Additional stream
		backbone (ResNet101) to utilize spatial information.
	}\label{tab:time}\vspace{-7pt}
	\begin{tabular}{l || c c c c }
		\toprule[1.5pt]
		Model		       &time(s) & FPS  &  param(M)\\
		\hline
		\hline
		Baseline 		  &0.029 &34 &56.8 \\
		Baseline+OG	      &0.033 &30 &57.7 \\
		Baseline+HHANet    &0.053 &18 &99.4 \\
		SGNet(ResNet50)  &0.028 &36 &39.3\\
		SGNet$^\dag$ &0.032 &31 &58.3\\
		SGNet  &0.037 &26 &58.3\\

		\bottomrule[1.5pt]
	\end{tabular}
\end{table}

\def\MthdN{Network & Backbone}
\def\FCN{FCN~\cite{fcn} & 2$\times$VGG16}
\def\LSDGF{LSD-GF~\cite{local} & 2$\times$VGG16}
\def\RefNt{RefineNet~\cite{refinenet} & ResNet152}
\def\ACNet{ACNet~\cite{hu2019acnet} & 2$\times$ResNet50}
\def\RDFNt{RDFNet~\cite{rdfnet} & 2$\times$ResNet152}
\def\RDFNtsmall{RDFNet~\cite{rdfnet} & 2$\times$ResNet101}
\def\CFNet{CFNet~\cite{cascaded} & 2$\times$ResNet152}
\def\DCNNs{D-CNN~\cite{dcnn} & 2$\times$VGG16}
\def\DCNNss{D-CNN~\cite{dcnn} & 2$ \times$ResNet152}
\def\DCNN{D-CNN~\cite{dcnn} & VGG16}
\def\DGNN{3DGNN~\cite{qi20173d} & VGG16}
\def\SGNetsmall{SGNet & ResNet50}
\def\SGNet{SGNet & ResNet101}
\def\SGNets{SGNet* & ResNet101}

\begin{table*}[t]\normalsize
	\centering
	\renewcommand\tabcolsep{12pt}
	\caption{\textbf{Comparison results on NYUDv2 test dataset.} 
	  MS: Multi-scale test;
		SI: Spatial information. The input image size for forward speed comparison is 
		$425 \times 560$ using NVIDIA 1080Ti following ~\cite{dcnn}. We add ASPP module~\cite{deeplab}
		after the final layer of SGNet, noted as``SGNet*".}
	\label{tab:nyud1} \vspace{-7pt}
	\begin{tabular}{l || c c c| c c c c c}
		\toprule[1.5pt]
		\MthdN &MS  &  SI & Acc & mAcc& mIoU& FPS& param (M)\\ \hline\hline
		\FCN   &    & HHA & 65.4 & 46.1 & 34.0 &  8 &272.2\\
		\LSDGF &    & HHA & 71.9 & 60.7 & 45.9 &  - &-    \\
		\DGNN  &		& HHA &  -   & 55.2 & 42.0 & 5  &47.2 \\
		\DCNN  & & Depth & - &53.6 & 41.0 & 26 & 47.0\\
		\DCNNss  &    &Depth&  -   & 61.1 & 48.4 & - &- \\ 
		\ACNet &    &Depth&  -   &  -   & 48.3 & 18 &116.6\\
		\RefNt &\Yes&  -  & 73.6 & 58.9 & 46.5 & 16 &129.5\\
		\RDFNt &\Yes& HHA & 76.0 & 62.8 & 50.1 & 9  &200.1\\
		\RDFNtsmall &\Yes& HHA & 75.6 & 62.2 & 49.1 & 11  &169.1\\
		\CFNet &\Yes& HHA &  -   &  -   & 47.7 & -  & -   \\
		\hline
		\SGNetsmall &	& Depth & 75.0 & 59.6 & 47.7 & \textbf{39} &\textbf{39.3}\\
		\SGNet &        & Depth & 75.6 & 61.9 & 49.6 & 28 &58.3\\
		\SGNets&       & Depth & 76.1 & 62.7 & 50.2&  26 &64.7\\
		\SGNets&\Yes    & Depth & \textbf{76.8}&\textbf{63.3}&\textbf{51.1}& 26 &64.7\\
		\bottomrule[1.5pt]
	\end{tabular}
\end{table*}

We can draw the following two conclusions from the results in the \tabref{tab:ablation1}.
1) The inference speed of the baseline network is fast, but its performance is poor.
Replacing convolution with S-Conv can improve the results of the baseline network
with a little bit more parameters and computational time.
2) In addition to the first convolution in layer 3 whose stride is 2,
the effect of replacing the later convolution is better.
The main reason would be that spatial information can better guide down-sampling operation in the first convolution.
%
Thus we choose to replace the first convolution and the last two convolutions
of each layer with S-Conv.
We generalize the rules found in layer 3 to other layers and achieve better results.
The above experiments show that our S-Conv can significantly improve network
performance with only a few parameters.
It is worth noting that our network has no spatial information stream.
The spatial information only affects the distribution and weight of
convolution kernel.
We also explore the performance of S-Conv embedded into different layers.
The results are shown in \tabref{tab:ablation_layer}.
We can observe that the performance enhances with the number of layers equipped with
S-Conv.

We also show the IoU improvement of S-Conv on most categories in
\figref{fig:catgory}. It's obvious that our S-Conv improves IoU in most categories,
especially for objects lacking representative texture information such as mirror, 
board and bathtub.
There are also clear improvements for objects with
rich spatial transformation, such as chairs and tables. 
This shows that our S-Conv can make good use of 
spatial information during the inference process.

\mypara{Architecture ablation}: To evaluate the effectiveness
of each component in our proposed S-Conv,
we design ablation studies.
The results are shown in \tabref{tab:ablation2}.
By default, we replace
the first convolution and the last two convolutions
of each layer according to~\tabref{tab:ablation1}.
We can see that the offset generator, spatial projection module, 
and weight generator
of S-Conv all contribute to the improvement of the results.

\mypara{Comparison with alternatives}: 
Most methods~\cite{fusenet,jiang2018rednet,rdfnet,hu2019acnet} 
use a two-stream network to extract features from two
different modalities and then combine them.
%
Our S-Conv focuses on advancing the feature extraction process
of the network by utilizing spatial information.
Here we compare our S-Conv with two-stream network, deformable convolution~\cite{deformablev2, deform},
and depth-aware convolution~\cite{dcnn}.
We use a simple baseline which consists of a ResNet101 
network with deep supervision and a simple decoder.
\begin{figure}[t]
	\centering
	\includegraphics[width = 1.0 \columnwidth]{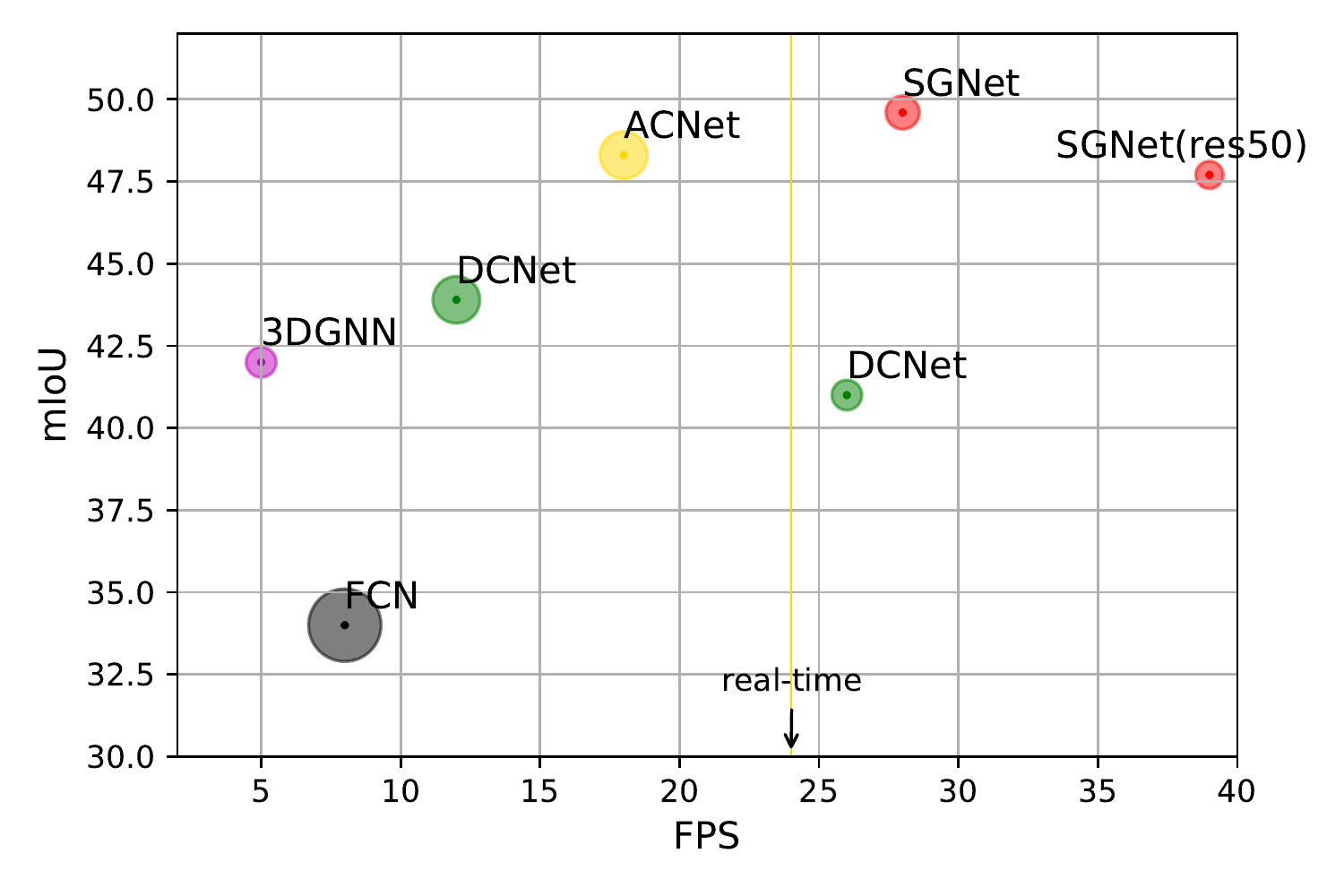}
	\caption{\textbf{FPS, mIoU, and the number of parameters 
	of different methods on NYUDv2.}
	    The input image size for all single-scale 
        speed comparisons
       is $425 \times 560$ following~\cite{dcnn}. 
		The radius of the circle corresponds to the number 
		of parameters of the model. 
		The results of DCNet~\cite{dcnn}
		and 3DGNN~\cite{qi20173d} are from ~\cite{dcnn}. 
		Our SGNet can achieve fastest inference time and state-of-the-art performance.
	}
	\label{fig:fps}
\end{figure} 
We add an additional ResNet101 network, called HHANet, 
to extract HHA features and fuse
it with our baseline features at the final layer 
of a two-stream network.
To compare with depth-aware convolution and deformable convolution, similar to SGNet, 
we replace the first convolution and the last two convolutions of each layer.
For ``Baseline + DAC + DCV2", we replace
convolution with depth-aware convolution~\cite{dcnn} (DAC) 
in first two layers and 
replace convolution with 
deformable  convolution~\cite{deform} (DCV2) in last two layers, 
because DCV2 does not work for the lower
layers~\cite{deform}.
The results are shown in \tabref{tab:promote}.
We find that our S-Conv achieves better results than two-stream networks, 
deformable convolution~\cite{deform},
depth-aware convolution~\cite{dcnn}, and their combination.
This demonstrates that our S-Conv can effectively utilizes spatial information.
The baseline equipped with weight generator can also achieve better results 
than depth-aware convolution,
indicating that learning weights from spatial information is necessary.
\begin{figure*}[htbp]
	\centering
	\begin{overpic}[width = 2\columnwidth]{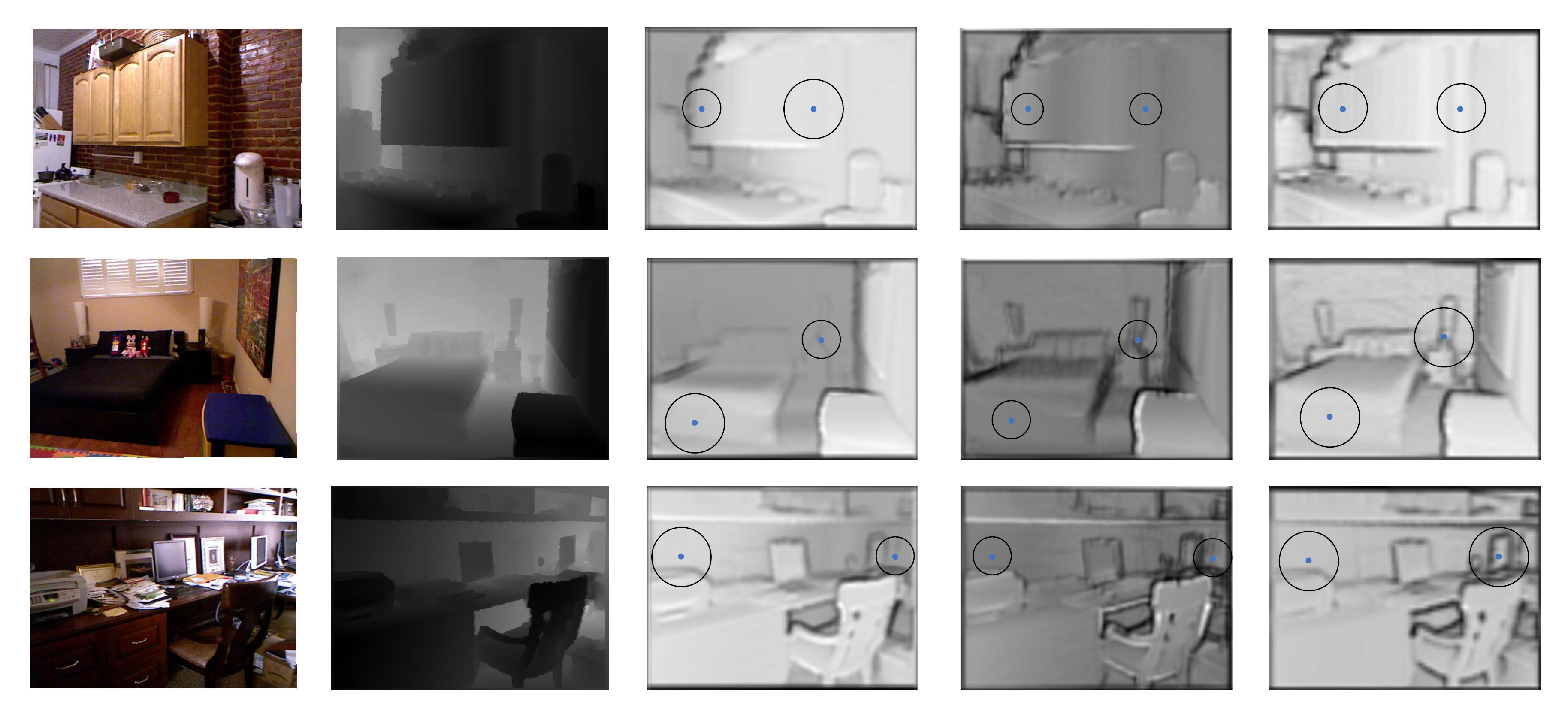}
		\put(7.5,0){RGB}
		\put(27,0){Depth}
		\put(47,0){Layer1\_1}
		\put(67,0){Layer1\_3}
		\put(86,0){Layer2\_1}
		\put(-1,7.5){(c)}
		\put(-1,23){(b)}
		\put(-1,38){(a)}
	\end{overpic}
	\caption{\textbf{The visualization of relative receptive field in S-Conv.}}
	\label{fig:visual}
\end{figure*}

\begin{figure*}[t]
	\centering
	\begin{overpic}[width = 2\columnwidth]{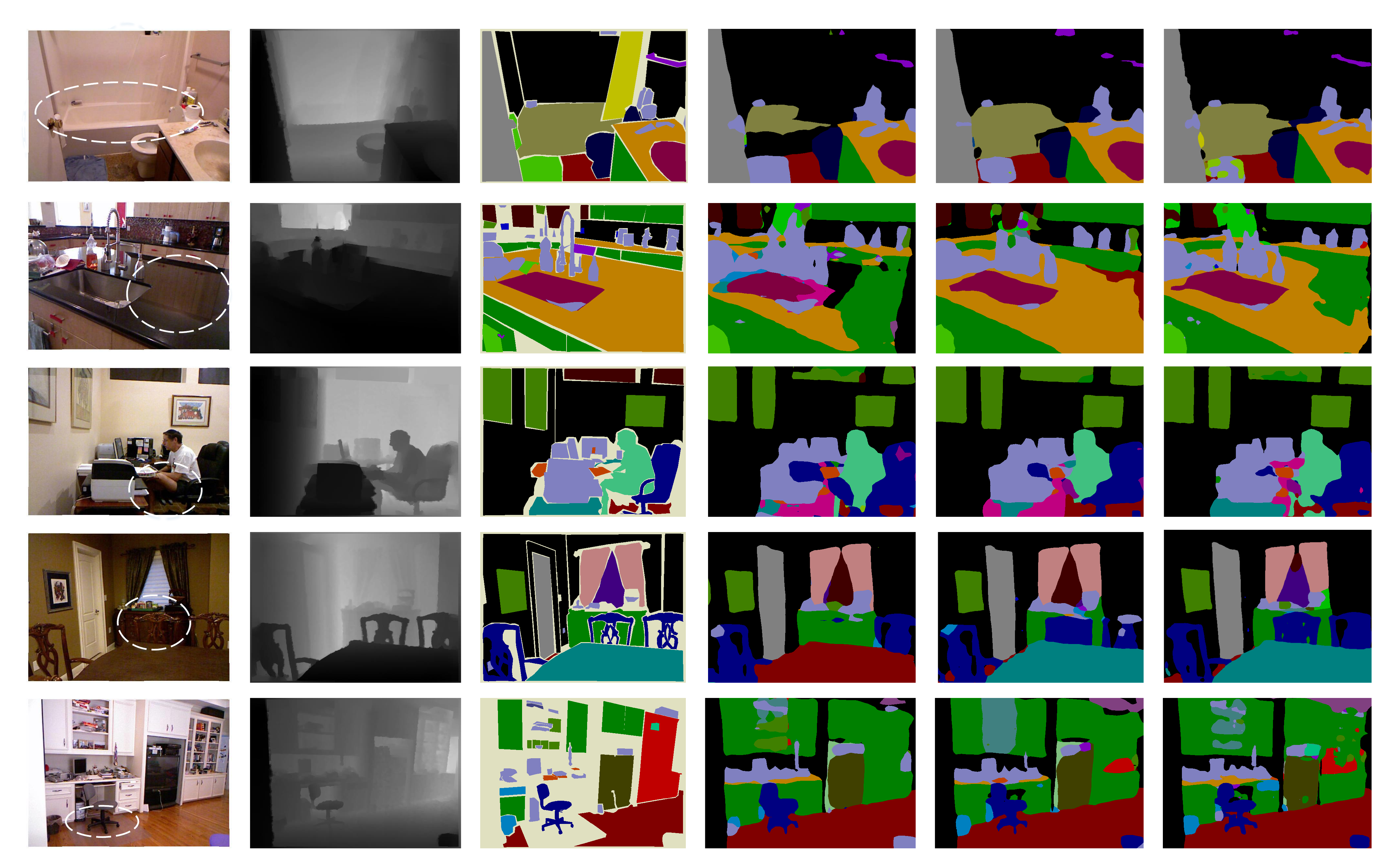}
		\put(7.5,0){RGB}
		\put(23,0){Depth}
		\put(40,0){GT}
		\put(55.5,0){Baseline}
		\put(72,0){SGNet}
		\put(88,0){SGNet-8s}
		\put(-1,6){(e)}
		\put(-1,18){(d)}
		\put(-1,30){(c)}
		\put(-1,43){(b)}
		\put(-1,55){(a)}
	\end{overpic}
	\caption{\textbf{The qualitative semantic segmentation comparison results 
	on NYUDv2 test dataset.} SGNet-8s: output stride is 8.}
	\label{fig:compare}
\end{figure*}
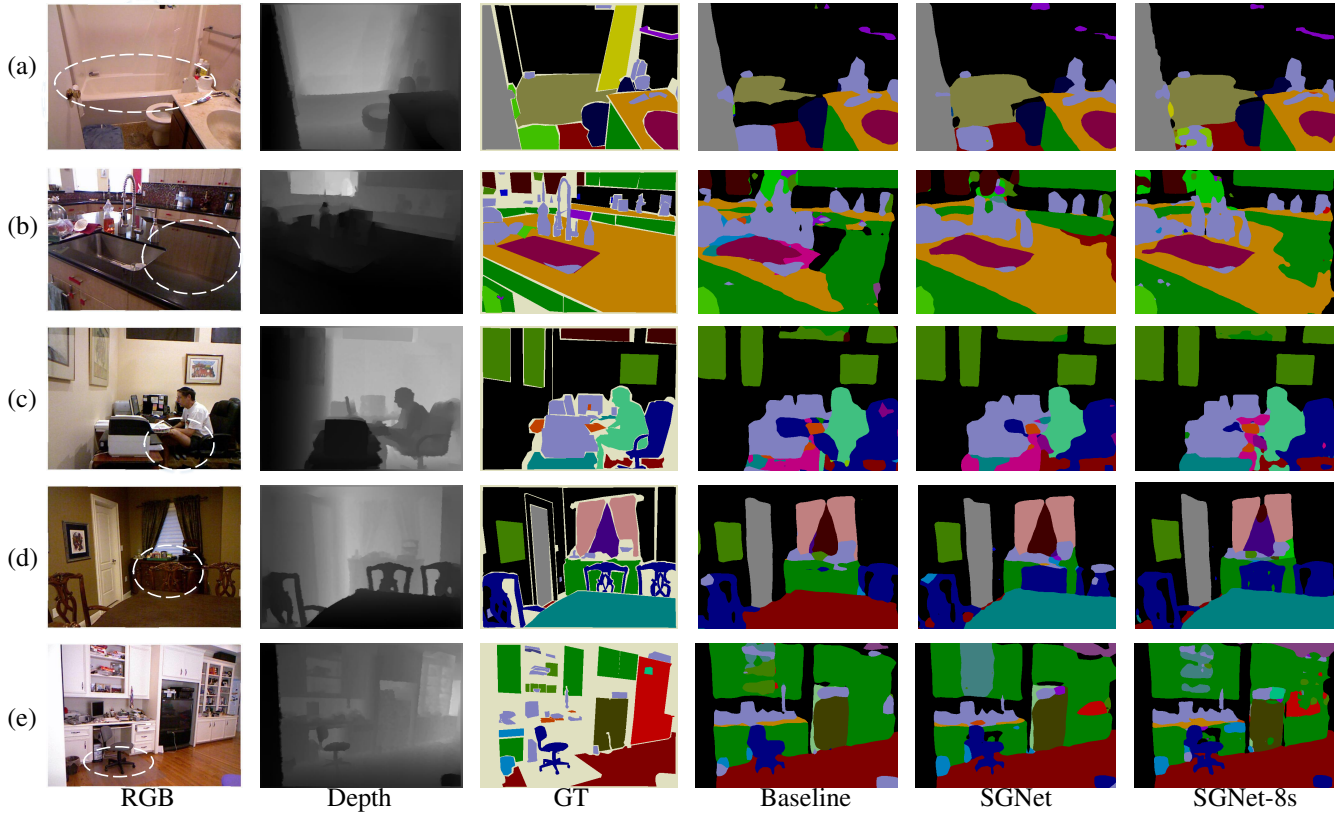

\begin{table*}[t]\normalsize
	\centering
	\caption{\textbf{Comparison results on SUNRGBD test dataset.} 
		MS: Multi-scale test, SI: Spatial information. 
		We add 
		ASPP module~\cite{deeplab} after the final layer of SGNet,
		noted as ``SGNet*".
	}\label{tab:sunrgbd}
	\vspace{-7pt}
	\renewcommand\tabcolsep{12.2pt}
	\begin{tabular}{l || c c c| c c c c} \toprule[1.5pt]
		\MthdN & MS & SI & Acc & mAcc& mIoU & param (M) \\ \hline \hline 
		\LSDGF &    &HHA & -    & 58.0 & -      & -     \\
		
		\RefNt &\Yes& -  & 80.6 & 58.5 & 45.9   & 129.5 \\
		CGBNet~\cite{ding2020semantic} & ResNet101   & &- & 82.3 &61.3 & 48.2 &-\\
		\DGNN  &\Yes&HHA & -    & 57.0 & 45.9   & 47.2  \\
		\DCNNs &    &HHA & -    & 53.5 & 42.0  & 92.0  \\
		\ACNet &    &HHA & -    & -    & 48.1   & 272.2 \\
		\RDFNt &\Yes&HHA & 81.5 & 60.1 & 47.7   & 200.1 \\
		\CFNet &\Yes&HHA & -    &  -   & 48.1   & -     \\
		\hline
		\SGNet &    &Depth & 81.0 & 59.6 & 47.1   & 58.3 \\
		\SGNets &    &Depth & 81.0 & 59.8 & 47.5   & 64.7 \\
		\SGNets &\Yes&Depth & 82.0 & 60.7 & \textbf{48.6}   & 64.7 \\
		\bottomrule[1.5pt]
	\end{tabular}
\end{table*}

\begin{figure}[htbp]
	\centering
	\begin{overpic}[width = 1\columnwidth]{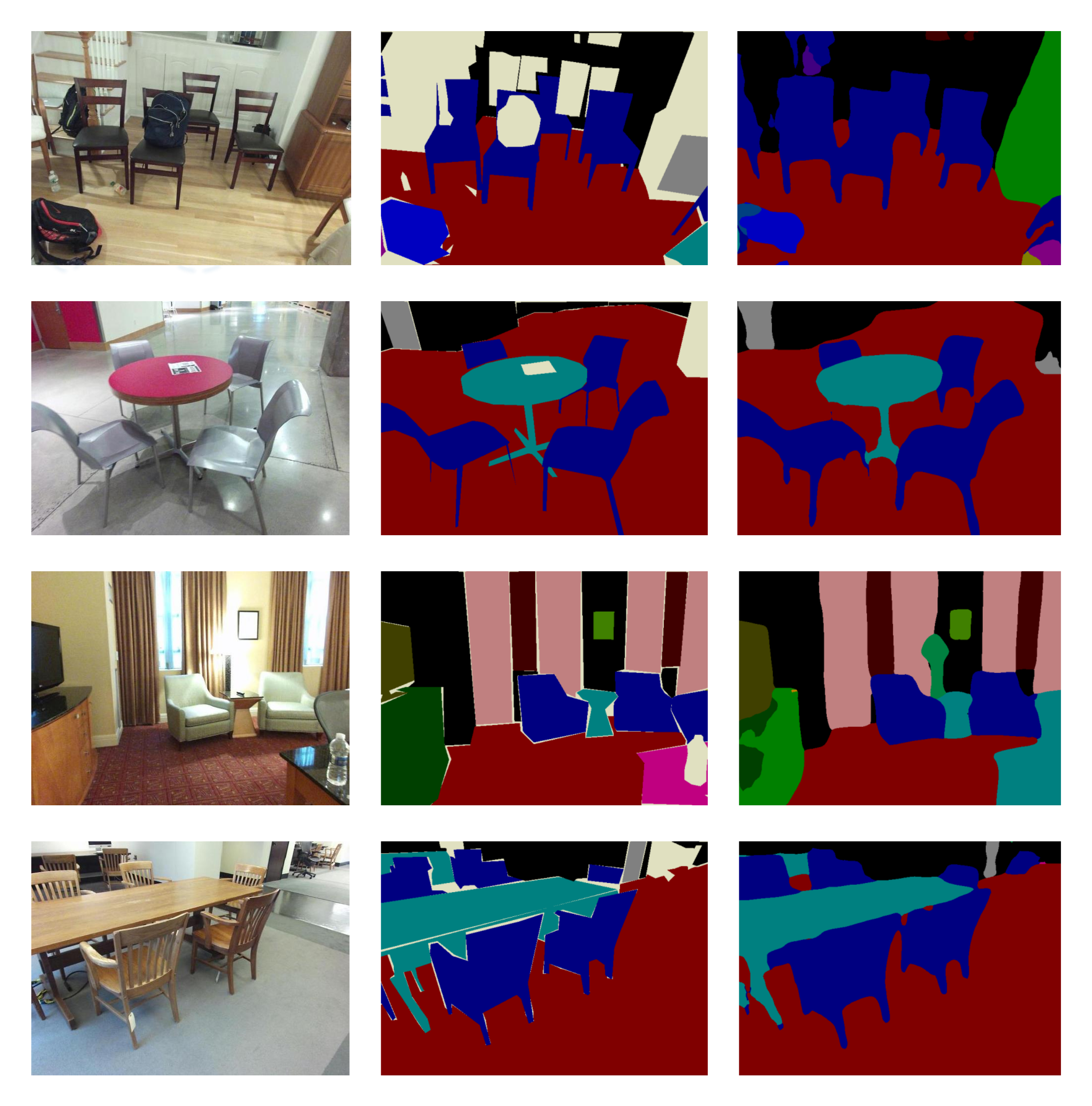}
		\put(14,0){RGB}
		\put(46,0){GT}
		\put(74,0){SGNet}
		\put(-2,12){(d)}
		\put(-2,36){(c)}
		\put(-2,60){(b)}
		\put(-2,84){(a)}
	\end{overpic}
	\caption{\textbf{The qualitative semantic segmentation comparison 
	results on SUNRGBD test dataset.}}
	\label{fig:sunrgbd}
\end{figure}

\mypara{Spatial information comparison}: 
%
We also evaluate the impact of different formats of spatial information on S-Conv.
The results are shown in \tabref{tab:ablation3}.
We can see that depth information leads to comparable
results with HHA and 3D coordinates, and better results than intermediate 
RGB features which are used by deformable convolution~\cite{deform,deformablev2}.
This shows the advantage of using spatial information for offset and weight generation over RGB features.
However, converting depth to HHA is time-consuming~\cite{fusenet}.
Hence 3D coordinates and depth map are more suitable for real-time segmentation using SGNet.
It can be seen that even without 
spatial information input (with RGB features), 
our S-Conv has more than 3.4\% improvement than the 
baseline.


\mypara{Inference speed test}: 
To demonstrate the light weight of S-Conv,
we test the inference speed of SGNet in this part.
We also compare our S-Conv with two-stream methods.
The input size of image is $480 \times 640$. Results are shown in \tabref{tab:time}.
We can observe that S-Conv only requires a small amount of additional computation compared with
two-stream methods. Our SGNet can also achieve real-time inference speed using ResNet101 and 
ResNet50~\cite{resnet} backbone. 

\subsection{Comparison with \sArt}
We compare our SGNet with other \sArt methods on
NYUDv2~\cite{nyud} and SUNRGBD~\cite{sunrgbd, sunrgbd2} datasets.
The architecture of SGNet is shown in \figref{fig:network}.
%

\mypara{NYUDv2 dataset:} The comparison results can be found in~\tabref{tab:nyud1}
and \figref{fig:fps}. We change the learning rate from 5e-3 to
8e-3. We down-sample the input image to $480 \times 640$ and
upsample its predict map to get final results during test.
To compare inference speed with other methods,
the input image size for all 
speed comparisons in~\tabref{tab:nyud1} 
is $425 \times 560$ following~\cite{dcnn}. 
The inference speed results of DCNet~\cite{dcnn}
and 3DGNN~\cite{qi20173d} are from ~\cite{dcnn},
and we test the single-scale speed of other methods
under the same conditions using NVIDIA 1080Ti in~\tabref{tab:nyud1}.
Furthermore, 
inference speed test of SGNet with input size $480 \times 640$ 
is shown in 
\tabref{tab:time}.
Note that some methods in \tabref{tab:nyud1} do not
report parameter quantities or open source.
So we just listed the mIoU of these methods.
We can draw the following conclusions from \tabref{tab:time} and
\tabref{tab:nyud1}.
Instead of using additional
networks to extract spatial features, 
our SGNet (ResNet50) can achieve
competitive performance and fastest inference with minimum number of 
parameters.
Our SGNet (ResNet101) can achieve
more competitive performance and real-time inference.
This benefits from S-Conv which can make use of spatial
information efficiently with only a small amount of extra 
parameters and computation cost.
Moreover, our S-Conv can achieve good results without using HHA information,
making it suitable for real-time tasks.
This verifies the efficiency of our S-Conv in
utilizing spatial information.
At the expense of a little bit more reasoning time by
adding ASPP module~\cite{deeplab} after SGNet noted as SGNet*,
the proposed SGNet can achieve better results than
other methods and 
RDFNet which uses multi-scale test, HHA information
and two ResNet152 backbones. After using multi-scale test
which is used by other methods,
SGNet's performance can be further improved.

%

\mypara{SUNRGBD dataset:}   The comparison results on the
SUNRGBD dataset are shown in \tabref{tab:sunrgbd}.
It is worth noting that some methods in \tabref{tab:nyud1}
did not report results
on the SUNRGBD dataset. The inference time and parameter number of 
models in \tabref{tab:sunrgbd}
are the same as those in \tabref{tab:nyud1}.
Our SGNet can achieve competitive results in real-time compared with 
models that do not have real-time performance.
SGNet's performance can be further improved 
by using multi-scale test.

\mypara{Cityscapes dataset:} 
We add ASPP~\cite{deeplab} module after SGNet and set $output \ stride = 8.$
noted as "SGNet-8s*".
We training with 2975 images on training set for validation.
We also provide our test result 
\footnote{\url{https://www.cityscapes-dataset.com/anonymous-results/?id=d772c1227bd7cfe7b841805796490cab82bb7c7749e5b1ec8e69e6e86134bfb3}} 
on Cityscapes server.
The comparison results on the
Cityscapes dataset are shown in \tabref{tab:cityscape}.
It is worth noting that due to the serious noise of depth map in Cityscapes, 
most of previous RGB-D based methods perform worse than RGB based methods.
We can observe that our network benefiting
from S-Conv can achieve better results than baseline and
achieve competitive results on Cityscapes.

\begin{table}[htbp]\normalsize
	\centering
	\caption{\textbf{Comparison results on Cityscapes validation dataset.
	\ddag: results on test dataset.}}
	\label{tab:cityscape} \vspace{-7pt}
	\renewcommand\tabcolsep{4.5pt}
	\begin{tabular}{c | c  c c l} \toprule[1.5pt]
		\MthdN & iterations & MS & mIoU \\ 
		\hline
		Baseline &ResNet101 & 40k & & 78.2\\
		SGNet-8s*  &ResNet101 & 40k &  & 79.2\\
		SGNet-8s*  &ResNet101 & 65k &\Yes & 80.6\\
		SGNet-8s*  &ResNet101 & 65k &\Yes & 81.2$^\ddag$\\
		\bottomrule[1.5pt]
	\end{tabular}
\end{table}

\subsection{Qualitative Performance}
\mypara{Visualization of receptive field in S-Conv}: Appropriate 
receptive field is very important for scene recognition. 
We visualize the input adaptive receptive field of SGNet in different layers generated by S-Conv. 
Specifically, we get the receptive field of each pixel by summing up the norm 
of their offsets during the S-Conv operation, 
then we normalize each value to [0, 255] and visualize the result using a gray-scale image. 
The results are shown in \figref{fig:visual}. 
The brighter the pixel, the larger the relative receptive field.
We also 
use the radius of circle to represent the size of the relative receptive field.
We observe that the receptive fields of different convolutions vary 
adaptively with the depth of the input image. 
For example, in layer1\_1, the receptive field is inversely 
proportional to the depth.
The combination of the adaptive receptive field 
learned at each layer can help the network better resolve indoor scenes with complex spatial relations.%

\mypara{Qualitative comparison results:}  
We show qualitative comparison results on NYUDv2 
test dataset in \figref{fig:compare}.
%
For the visual results in \figref{fig:compare}(a),
the bathtub and the wall have insufficient texture, which cannot 
be easily distinguished by the baseline method.
Some objects may have reflections such as the table in \figref{fig:compare}(b),
which is also challenging for the baseline.
SGNet, however, can recognize it well by incorporating 
spatial information with the help of S-Conv.
%
%
The chairs in \figref{fig:compare}(c, d) 
are hard to be recognized by RGB data due to the low contrast and
confused texture,
while they can be easily recovered by SGNet benefiting 
from the equipped S-Conv.
In the meantime, SGNet can recover the object's geometric shape nicely,
as demonstrated by the chairs of \figref{fig:compare}(e).
We also show qualitative results on SUNRGBD 
test dataset in \figref{fig:sunrgbd}. 
It can be seen that our SGNet can also achieve precise segmentation on SUNRGBD.



\section{Conclusion}
\label{sec:conclution}
In this paper, we propose a novel \emph{\textbf{S}patial information guided 
	\textbf{Conv}olution (S-Conv)}
operator.
Compared with conventional 2D convolution, it can adaptively
adjust the convolution weights and distributions according
to the input spatial information, resulting in better awareness
of the geometric structure with only a few additional parameters and computation cost.
We also propose \textbf{S}patial information \textbf{G}uided convolutional 
\textbf{Net}work (SGNet) equipped with S-Conv that yields real-time
inference speed and achieves competitive results on NYUDv2 and
SUNRGBD datasets for RGBD semantic segmentation.
We also compare the performance of using different inputs to generate offset, 
demonstrating the advantage of using spatial information over RGB feature.
Furthermore, we visualize the depth-adaptive receptive field in each layer to show effectiveness.
In the future, we will investigate the fusion of different modal information 
and the adaptive change of S-Conv structure simultaneously,
making these two approaches benefit each other. 
We will also explore the application of S-Conv in different fields,
such as pose estimation and 3D object detection.
%


%

%
%
%
\section*{Acknowledgment}
This research was supported by Major Project for New
Generation of AI under Grant No. 2018AAA0100400,
NSFC (61620106008), Tianjin Natural Science Foundation (17JCJQJC43700),
and S\&T innovation project from Chinese Ministry of Education.

\ifCLASSOPTIONcaptionsoff
  \newpage
\fi



%
\bibliographystyle{IEEEtran}
\bibliography{egbib}

\vfill

\end{document}